%% file: Template.tex
\documentclass{article}
\usepackage{spconf,amsmath,graphicx}
\usepackage{standalone}

\usepackage{multirow}
\usepackage{array}
\usepackage{siunitx}

\usepackage{bbm}
\usepackage{xcolor}
\usepackage{tikz}
\usetikzlibrary{positioning, arrows.meta, calc, shadows, backgrounds, shapes.geometric, fit}
\usepackage{bm}
\usepackage{eso-pic}

\definecolor{bitgray}{RGB}{128, 128, 128}
\definecolor{containergray}{RGB}{220, 220, 220}

\newcommand{\ml}[1]{\begin{tabular}{@{}l@{}}#1\end{tabular}}

\title{TAFA-GSGC: Group-wise Scalable Point Cloud Geometry Compression with Progressive Residual Refinement}
\name{Author(s) Name(s)\thanks{Thanks to XYZ agency for funding.}}
\address{Author Affiliation(s)}

\name{Xiumei Li, Alexander Kopte and Andr\'e Kaup  \thanks{This work was funded by the Deutsche Forschungsgemeinschaft (DFG, German Research Foundation) under Grant SFB 1483 – Project-ID 442419336.}}
\address{Chair of Multimedia Communications and Signal Processing\\
Friedrich-Alexander-Universit\"at Erlangen-N\"urnberg (FAU)\\
Erlangen, Germany}
\begin{document}
\ninept

\AddToShipoutPictureBG*{%
  \AtPageUpperLeft{%
    \raisebox{-3\baselineskip}{% Adjusts vertical position from the top of the page
      \makebox[\paperwidth][c]{% Creates a box the full width of the paper and centers the content
        \parbox{0.9\textwidth}{\centering\footnotesize % The notice itself, in a slightly narrower box
        % This work has been submitted to the IEEE for possible publication. Copyright may be transferred without notice, after which this version may no longer be accessible.
          © 2026 IEEE. Personal use of this material is permitted. Permission from IEEE must be obtained for all other uses, in any current or future media, including reprinting/republishing this material for advertising or promotional purposes, creating new collective works, for resale or redistribution to servers or lists, or reuse of any copyrighted component of this work in other works.
        }%
      }%
    }%
  }%
}

\maketitle
\begin{abstract}

Scalable compression is essential for bandwidth-adaptive transmission, yet most learned codecs are optimized for a fixed rate-distortion point, making rate adaptation costly due to re-encoding or maintaining multiple bitstreams. In this work, we propose \emph{TAFA-GSGC}, a scalable learned point cloud geometry codec that enables multi-quality decoding from a \emph{single} bitstream and a \emph{single} trained model. TAFA-GSGC combines layered residual refinement with channel-group entropy coding, and introduces a \emph{Target-Aligned Feature Aggregation} module to reduce cross-layer redundancy in enhancement residuals. Our framework supports up to \emph{9} decodable quality levels with monotonic quality improvement as more sub-bitstreams are received, while maintaining strong compression efficiency. Compared with the PCGCv2 baseline, TAFA-GSGC demonstrates improved RD performance, achieving average BD-rate reductions of  \mbox{4.99\%} and \mbox{5.92\%} in terms of D1-PSNR and D2-PSNR, respectively.

% achieving \mbox{-4.99\%/-5.92\%} average BD-Rate savings (D1/D2). 

% Under a fair comparison, TAFA-GSGC attains comparable or slightly better RD performance than the learned baseline PCGCv2, achieving \mbox{-4.99\%/-5.92\%} average BD-Rate savings (D1/D2).

% The abstract should appear at the top of the left-hand column of text, about
% 0.5 inch (12 mm) below the title area and no more than 3.125 inches (80 mm) in
% length.  Leave a 0.5 inch (12 mm) space between the end of the abstract and the
% beginning of the main text.  The abstract should contain about 100 to 150
% words, and should be identical to the abstract text submitted electronically
% along with the paper cover sheet.  All manuscripts must be in English, printed
% in black ink.
\end{abstract}

\section{Introduction}
\label{sec:intro}

Real-world point clouds often contain millions of sparsely and irregularly sampled 3D points, making raw storage and transmission expensive and compression inherently challenging~\cite{quach2022survey,wang2022sparse}.

Given these characteristics, MPEG~\cite{schwarz2018emerging} has standardized point cloud coding solutions, including G-PCC for static point clouds and V-PCC for dynamic point clouds. 
G-PCC typically exploits 3D structural representations for geometry coding, whereas V-PCC projects point clouds onto 2D patches and leverages video coding tools for compression. 
However, relying on hand-crafted modules, these conventional approaches often have difficulty capturing sparse and irregular geometry, leading to noticeable loss of fine structures at low bitrates~\cite{cao2021compression}.

In recent years, learned compression has become increasingly competitive for both images and point clouds, typically adopting an end-to-end autoencoder with entropy modeling and rate--distortion (RD) optimization~\cite{10916627,11175396}. 
Although learned geometry codecs can significantly improve RD performance over conventional methods, most existing approaches are non-scalable: different bitrate-quality operating points often require training multiple rate-specific models and storing or re-encoding separate bitstreams, which increases deployment cost and limits flexibility. 
In practical systems, however, available bandwidth, device compute, and target quality can vary over time, making dynamic rate and quality adaptation essential.

To address these issues, we propose a learned scalable lossy point cloud geometry codec built upon PCGCv2~\cite{wang2021multiscale}. 
Our method produces a single embedded bitstream that can be progressively truncated to decode multiple quality levels, enabling flexible rate control without retraining or maintaining multiple models. 
To improve compression efficiency, we introduce a Target-Aligned Feature Aggregation (TAFA) module that extracts target-aligned residual information for the enhancement layer. 
Moreover, we enable channel-wise scalability by partitioning latent channels into groups and entropy-coding them independently, which provides finer-grained rate adaptation and wider bitrate coverage.

\section{Related Work}
\label{sec:format}

MPEG standardizes point cloud compression through G-PCC and V-PCC~\cite{graziosi2020overview}. 
G-PCC uses octree-based occupancy signaling, with a Trisoup mode that approximates surfaces 
using triangle primitives, which can be advantageous at low bitrates. In contrast, V-PCC 
adopts a projection-based approach, packing point clouds into 2D patch atlases and compressing 
them with standard video codecs like HEVC or VVC~\cite{li2024mpeg}.

Learned point cloud geometry compression has advanced rapidly in recent years. 
Among them, PCGCv2~\cite{wang2021multiscale} is a representative sparse-tensor autoencoder framework that employs multiscale processing with progressive down-/up-sampling, and integrates learned transforms with octree-based geometry coding in its pipeline. 
Building on this line of work, SparsePCGC~\cite{wang2022sparse} improves compression efficiency by combining multiscale sparse-tensor representations with probabilistic occupancy modeling to better exploit point cloud sparsity and cross-scale dependencies. 
UniPCGC~\cite{10.1609/aaai.v39i12.33387} further targets practical deployment via a unified and efficient design, aiming for robust performance across a wide range of operating points. 
Collectively, these methods demonstrate strong RD performance on standard benchmarks, but they typically do not provide scalability from a single bitstream, motivating the scalable designs reviewed next.

% \begin{figure*}[t]
%     \centering

%     \begin{minipage}[c]{0.78\textwidth}
%         \centering
%         \includestandalone[width=0.98\textwidth]{overview}
%         % \caption{Scalable geometry codec.}
%         \caption{Framework of TAFA-GSGC.}
%         \label{fig:overview_codec}
%     \end{minipage}
%     \begin{tikzpicture}[baseline=0]
%         \draw[dotted, thick, gray] (0,-3cm) -- (0, 3.125cm);
%     \end{tikzpicture}
%     \hfill
%     \begin{minipage}[c]{0.2\textwidth}
%         \centering
%         \includestandalone[width=1\linewidth]{tafa}
%         % \captionsetup{skip=20pt}
%         \vspace{26pt}
%         \caption{TAFA module.}
%         \label{fig:overview_tafa}
%     \end{minipage}

%     \caption{Overview of TAFA-GSGC : DS/US (downsampling/upsampling), 
% TAFA (Target-Aligned Feature Aggregation), SConv (sparse convolution), Q (quantization), 
% AE/AD (arithmetic encoder/decoder), OE/OD (octree encoder/decoder), C (channel concatenation), 
% CLS (occupancy classification head).}
%     \label{fig:overview}
% \end{figure*}

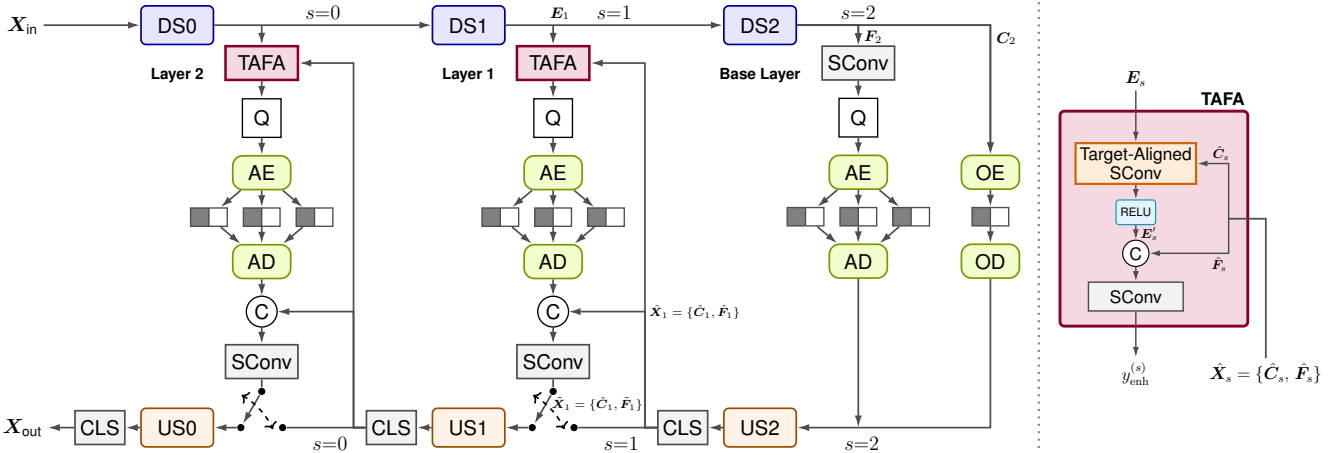
\begin{figure*}[t]
    \centering
    \begin{minipage}[c]{0.78\textwidth}
        \centering
        \includestandalone[width=0.98\textwidth]{overview}
    \end{minipage}
    \begin{tikzpicture}[baseline=0]
        \draw[dotted, thick, gray] (0,-2.85cm) -- (0, 3.125cm);
    \end{tikzpicture}
    \hfill
    \begin{minipage}[c]{0.2\textwidth}
        \centering
        \includestandalone[width=1\linewidth]{tafa}
    \end{minipage}
    \caption{Overview of our TAFA-GSGC (\textbf{left}) and TAFA module (\textbf{right}). DS/US, SConv, Q, AE/AD, OE/OD, Prune, and C represent downsampling/upsampling blocks, sparse convolution, quantization, arithmetic encoder/decoder,  octree encoder/decoder, pruning, and channel concatenation, respectively.}
    \label{fig:overview}
\end{figure*}

Recent works have started exploring learned scalable point cloud codecs. 
DL-PCSC~\cite{guarda2020point} is an early approach that enables quality scalability by splitting the latent representation of a dense 3D convolutional autoencoder into independently entropy-coded channel groups, forming a progressively decodable bitstream. While it improves RD performance over MPEG G-PCC on static geometry, it relies on independent latent-group coding without exploiting inter-layer dependencies to reduce cross-layer redundancy.

Mari \textit{et al.}~\cite{mari2024point} proposed SQH, which adds a quality-conditioned latent probability estimator QuLPE on top of JPEG PCC~\cite{10916627} to exploit cross-quality latent correlation, using decoded low-quality latents as side information to entropy-code higher-quality enhancement latents more efficiently. SQH achieves quality scalability with minor RD overhead. However, it remains tied to multiple pre-trained JPEG PCC models to generate target-quality latents, and thus does not remove the multi-model deployment cost; moreover, scalability is restricted to these discrete operating points and depends on JPEG PCC-specific training-induced latent alignment.

Pang \textit{et al.}~\cite{pang2022grasp} propose GRASP-Net, a two-layer lossy point cloud geometry codec with a G-PCC octree base layer and a learned enhancement layer that encodes local geometric residuals via point-based feature extraction and sparse-convolutional analysis/synthesis transforms. While GRASP-Net improves RD performance over G-PCC on both dense and sparse point clouds, its scalability is limited to a binary base/enhancement switch rather than multi-level quality adaptation.
 
Nguyen \textit{et al.}~\cite{nguyen2023deep} propose MNeT, an end-to-end multi-scale framework for \emph{lossless} point cloud attribute compression with a scalable mode. It uses a hierarchical probabilistic model for arithmetic coding, avoiding point-wise autoregressive prediction and achieving competitive lossless bitrates against G-PCC anchors. However, its scalability is mainly realized via reconstructions at different scales, rather than fine-grained truncation-based quality points from a single bitstream, which may limit rate-control precision.

Hu et al.~\cite{hu2025bits} applies scalable coding to 3D Gaussian splatting by leveraging an octree-based level of detail (LoD) hierarchy, where truncating the bitstream enables progressive decoding at different octree levels and the decoded features are used to predict Gaussian parameters for rendering. However, its scalability is largely limited to a few octree-level LoD points with relatively coarse granularity, rather than dense fine-grained rate adaptation.

Different from these approaches, our work emphasizes fine-grained scalability: by truncating a single bitstream, a single trained model can decode point clouds at up to 9 distinct quality levels, while maintaining strong compression efficiency. Moreover, our layered architecture and channel-group-wise design provide a natural path to supporting more decodable levels, thereby accommodating diverse application requirements.

\section{Scalable Framework}
\label{sec:scalable_framework}

Motivated by PCGCv2, we propose TAFA-GSGC, a group-wise scalable point cloud geometry compression framework with progressive residual refinement (Fig.~\ref{fig:overview}, left).
The framework consists of a base layer, two enhancement layers, and a channel-wise scalability mechanism.
The base layer adopts the backbone of PCGCv2 to produce a coarse reconstruction together with multi-scale representations.
Building on this reconstruction, each enhancement layer encodes the remaining geometric details as a residual representation, enabling progressive coarse-to-fine refinement. Beyond the layer-wise residual refinement, we further increase the number of decodable levels by partitioning the latent channels at each layer into multiple channel groups.
Each group is entropy-coded into an individual sub-bitstream, and all sub-bitstreams are concatenated in a deterministic order to form a single bitstream.
This structure enables rate adaptation by truncating the bitstream at sub-bitstream boundaries: decoding only the first $L$ sub-bitstreams yields a corresponding reconstruction quality, while higher $L$ progressively refines the reconstructed point cloud.
In contrast to PCGCv2, which targets a fixed single-quality operating point, TAFA-GSGC supports bitrate-adaptive transmission and multi-quality decoding from a single bitstream without re-encoding.

\subsection{Base layer}
\label{ssec:baselayer}

Our base layer follows the multiscale encoder--decoder backbone of PCGCv2.
Given an input point cloud $\bm{X}_{\text{in}}$, we voxelize it into a sparse tensor $\{\mathbf{C}_X,\mathbf{F}_X\}$, where $\mathbf{C}_X$ contains the coordinates of positively occupied voxels and $\mathbf{F}_X$ is initialized as an all-ones vector to indicate occupancy at these coordinates.

On the encoder side, the input is processed by a cascade of three downsampling blocks ($\mathrm{DS0}$--$\mathrm{DS2}$).
Each block first extracts local geometric patterns with sparse convolutions and then downsamples the spatial resolution by a factor of $2$ in all three dimensions, yielding a compact bottleneck representation $\{\mathbf{C}_{2},\mathbf{F}_{2}\}$.
At the bottleneck, the latent representation consists of (i) the set of occupied coordinates $\mathbf{C}_{2}$ and (ii) the associated latent features $\mathbf{F}_{2}$. $\mathbf{C}_{2}$ is losslessly coded using an octree-based geometry codec, while $\mathbf{F}_{2}$ is further encoded by our channel-wise scalable entropy coding scheme.

On the decoder side, a symmetric upsampling path ($\mathrm{US2}$--$\mathrm{US0}$) reconstructs geometry in a coarse-to-fine manner. 
After each upsampling stage, an occupancy classification head ($\mathrm{Prune}$) predicts voxel occupancy probabilities for each generated voxel; we retain only the top-$K$ voxels ranked by occupancy probability, discarding low-probability voxels and keeping the sparse tensor compact to reduce subsequent computation. The multi-scale reconstructions produced by the base layer are further used as conditioning signals for the enhancement layers to align and encode the missing residual details. Specifically, the coarsest-scale reconstruction from the classification head after $\mathrm{US2}$ is fed to the Layer~1 as its conditioning input.
We describe the enhancement layers and their conditional residual coding in Sec.~\ref{ssec:Enhancementlayer}.

\subsection{Enhancement Layers}
\label{ssec:Enhancementlayer}

The enhancement layers aim to transmit only the geometric details missing from the current reconstruction with a small additional bit\-rate, enabling noticeable quality gains on top of the base layer.
The key challenge is to avoid re-encoding information that is already predictable given the current reconstruction, i.e., to represent residual details with minimal cross-layer redundancy.

Prior scalable approaches~\cite{pang2022grasp,xie2024roi} quantize the point cloud geometry 
to form a base-layer reconstruction, which is encoded with an octree-based coder. The 
enhancement layer is then constructed by computing voxel-wise residuals between the input 
and the dequantized base reconstruction. However, this pipeline relies solely on fixed 
quantization and geometry coding, lacking learnable modules to strengthen the base layer. 
As a result, the base reconstruction has limited representational capacity and base-layer quality becomes a performance bottleneck.

To address this limitation, we propose TAFA, which fuses domain-aware features with reconstruction-aligned information to model residual details conditioned on the current reconstruction, reducing redundancy and improving coding efficiency. In the following, we present the architecture of TAFA and explain how it is integrated into the enhancement layers for residual coding.

\subsubsection{Target-Aligned Feature Aggregation (TAFA)}
\label{sssec:Target-AlignedFeatureAggregation}

TAFA is a cross-coordinate-set feature aggregation module, as illustrated in Fig.~\ref{fig:overview} (right).
Consider an enhancement layer operating at scale $s$.
Let $\mathbf{E}_s=\{\mathbf{C}_s,\mathbf{F}_s\}$ denote the encoder-side sparse tensor at scale $s$ (computed from the input), and let $\hat{\mathbf{X}}_s=\{\hat{\mathbf{C}}_s,\hat{\mathbf{F}}_s\}$ be the reconstructed sparse tensor at the same scale decoded from the previous layer.
TAFA aggregates encoder features ${\mathbf{F}}_s$ onto the decoder's coordinate set $\hat{\mathbf{C}}_s$. Concretely, $\mathbf{E}'_s$ is computed by applying a sparse convolution to $\mathbf{E}_s$, with its output coordinates explicitly specified as $\hat{\mathbf{C}}_s$, such that each output feature at a target coordinate in $\hat{\mathbf{C}}_s$ is obtained from encoder-side features $\mathbf{F}_s$. This enables encoder-side information to refine incomplete decoder representations during progressive residual reconstruction. 

% Through encoder-to-decoder cross-coordinate-set feature aggregation, TAFA computes $\mathbf{E}'_s$ on the target coordinate set $\hat{\mathbf{C}}_s$ by aggregating local neighborhoods from $\mathbf{E}_s$.

% \begin{equation}
% \mathbf{E}'_s = \mathrm{TAFA}(\mathbf{E}_s, \hat{\mathbf{C}}_s),
% \label{eq:tafa}
% \end{equation}

We then concatenate the aligned encoder context $\mathbf{E}'_s$ with the current reconstruction features $\hat{\mathbf{F}}_s$ along the channel dimension and project them to the enhancement latent space:

\begin{equation}
\mathbf{y}^{(s)}_{\mathrm{enh}} = g^{(s)}_{\mathrm{proj}}\Big(\big[\mathbf{E}'_s \, \| \, \hat{\mathbf{F}}_s\big]\Big),
\label{eq:enh_latent}
\end{equation}
where $g^{(s)}_{\mathrm{proj}}(\cdot)$ is implemented by a sparse convolution.
The resulting latent $\mathbf{y}^{(s)}_{\mathrm{enh}}$ is then encoded and transmitted using the proposed channel-wise scalable entropy coding scheme (discussed in Sec.~\ref{sec:channel_scalability}).
Since $\mathbf{y}^{(s)}_{\mathrm{enh}}$ is generated conditioned on the current reconstruction, it tends to suppress predictable components and allocate bits to missing details.

\subsubsection{Residual Fusion}
\label{sssec:Residualfusion}
At the decoder, the enhancement latent is decoded and fused with the previous-layer reconstruction $\hat{\mathbf{X}}_s=\{\hat{\mathbf{C}}_s,\hat{\mathbf{F}}_s\}$ at the same scale via channel concatenation and a sparse convolution, yielding a refined sparse tensor $\tilde{\mathbf{X}}_s=\{\hat{\mathbf{C}}_s,\tilde{\mathbf{F}}_s\}$.
The resulting tensor $\tilde{\mathbf{X}}_s$ incorporates the decoded enhancement information on the reconstructed voxels and is fed into the subsequent upsampling blocks to further refine the reconstruction.

In this work, we design two enhancement layers with identical architecture. They operate at scales $s\in\{0,1\}$ to refine the intermediate and the finest resolutions, respectively.
Concretely, Layer~1 encodes the residual details with respect to the base-layer reconstruction after $\mathrm{US2}$ (at scale $s=1$), while Layer~2 encodes the residual details with respect to the reconstruction produced after Layer~1 followed by $\mathrm{US1}$ (at scale $s=0$).

\subsection{Channel-Wise Scalability}
\label{sec:channel_scalability}

Inspired by~\cite{guarda2020point}, to enable fine-grained rate adaptation, we introduce channel-wise scalability for the latent representation.
We organize scalability in two dimensions: (i) \emph{across layers} {(Base Layer $\rightarrow$ Layer~1 $\rightarrow$ Layer~2)} and (ii) \emph{within each layer} via channel groups.
For each layer $\ell \in \{{\text{Base Layer}, \text{Layer 1}, \text{Layer 2}\}}$, we partition the latent representation $\mathbf{y}_{\ell}$ into $M_\ell$ channel groups and entropy-code each group independently, producing one sub-bitstream per group.
All sub-bitstreams are serialized in a fixed order, with Base-Layer groups followed by Layer~1 groups and then Layer~2 groups.
% \begin{equation}
% \mathcal{B} = [\mathcal{B}^{(\text{Base})}, \mathcal{B}^{(\text{L1})}, \mathcal{B}^{(\text{L2})}], \quad
% \mathcal{B}^{(\ell)} = [\mathbf{b}_1^{(\ell)}, \ldots, \mathbf{b}_{M_{\ell}}^{(\ell)}],
% \end{equation}
\begin{equation}
\mathcal{B} = [\mathcal{B}_{\text{Base}}, \mathcal{B}_{\text{L1}}, \mathcal{B}_{\text{L2}}], \quad
\mathcal{B}_{\ell} = [\mathbf{b}_{1,\ell}, \ldots, \mathbf{b}_{M_{\ell},\ell}],
\end{equation}
where each $\mathbf{b}_{m,\ell}$ forms a separate sub-bitstream segment.

At the decoder, we select a decode level $L$ and decode the first $L$ sub-bitstreams in the predefined order.
This yields a partial set of decoded channel groups across the three layers, while all remaining (undecoded) channel groups are filled with zeros, so that the latent tensors in each layer keep a fixed shape and can be fed into the corresponding synthesis modules without ambiguity.
As $L$ increases, more channel groups become available and the reconstruction is progressively refined.

When only Base Layer groups are decoded, the decoder reconstructs a coarse geometry from the Base Layer latent and directly synthesizes the point cloud through the upsampling path (US2$\rightarrow$US1$\rightarrow$US0).
Layer~1 transmits residual geometry information that is conditioned on the current reconstruction: the intermediate output produced at US2 serves as the conditioning signal for TAFA of Layer~1 to extract residual features, which are then fused with the decoded Layer~1 residual latent before being propagated through the subsequent upsampling blocks to refine the reconstruction.
Layer~2 follows the same residual-refinement pipeline, but uses the refined reconstruction from Layer~1 as its conditioning input (e.g., the output after US1), enabling further correction of the remaining geometric details and producing the finest output.

\subsection{Loss Function}
\label{ssec:LossFunction}

%-------------1
We train the proposed scalable codec using a RD loss. Given a target decode level $L$, the loss is
\begin{equation}
\mathcal{L}(L)= R_{\text{total}}(L) + \lambda(L)\sum_{i=1}^{S} D_i(L),
\label{eq:rd_obj}
\end{equation}
where $S$ denotes the number of scales, $R_{\text{total}}(L)$ is the total rate of the latent groups decoded up to level $L$, and $D_i(L)$ is the distortion measured at scale $i$.
$\lambda(L)$ is a level-dependent Lagrange multiplier that controls the RD trade-off for the operating point selected by the decode level $L$.
In our design, each sub-bitstream (i.e., each channel group) corresponds to one decodable level. A distinct multiplier is assigned to each level, enabling control over the overall bitrate range by adjusting these multipliers.
Therefore, the number of Lagrange multipliers equals the total number of channel groups across the Base Layer, Layer~1, and Layer~2, i.e., $M_{\text{base}} + M_{\text{layer1}} + M_{\text{layer2}}$.

Let $\hat{\mathbf{z}}(L)$ denote the set of quantized latent symbols from the channel groups decoded up to level $L$.
Given the likelihoods $p(\hat{z}_k)$ predicted by the learned entropy model, the rate is estimated by
\begin{equation}
R_{\text{total}}(L)=\frac{1}{N_{\text{in}}}\sum_{\hat{z}_k\in\hat{\mathbf{z}}(L)} -\log_2 p(\hat{z}_k),
\label{eq:rate}
\end{equation}
where $N_{\text{in}}$ is the number of input points for normalization.

For the distortion term, we adopt a multiscale occupancy classification loss from PCGCv2.
At scale $i\in\{0,\ldots,S\}$, the decoder predicts occupancy probabilities $p_v\in(0,1)$ for voxels $v\in\mathcal{V}_i$, 
where $\mathcal{V}_i$ denotes the generated voxel set at scale $i$.
The ground-truth occupied voxel set at the same scale is denoted by $\mathcal{G}_i$ (from encoder), 
and the corresponding binary target is defined as $y_v=1[v\in\mathcal{G}_i]$.
We then compute the average binary cross-entropy (BCE) over $\mathcal{V}_i$ at decode level $L$:
% \begin{equation}
% \begin{split}
% D_i(L)=\frac{1}{|\mathcal{V}_i(L)|}\sum_{v\in\mathcal{V}_i(L)}
% \Big(&-y_v\log(p_v(L)) \\
% &-(1-y_v)\log\!\big(1-p_v(L)\big)\Big).
% \end{split}
% \end{equation}

\begin{equation}
\begin{split}
D_i(L)=\frac{1}{|\mathcal{V}_i(L)|}\sum_{v\in\mathcal{V}_i(L)}
\Big(&-y_v\log(p_v(L)) \\
&-(1-y_v)\log\!\big(1-p_v(L)\big)\Big).
\end{split}
\end{equation}

\section{Experimental Setup}
\label{sec:ExperimentalSetup}

% ---------------------need to modify--------
%Dataset
Since our method builds upon PCGCv2 and augments it with a scalable coding mechanism, we follow the same training data and preprocessing pipeline as PCGCv2 to enable a controlled and reproducible comparison. Specifically, the training set is constructed from dense point clouds sampled from ShapeNet~\cite{chang2015shapenet} containing approximately 26{,}000 models. We split the data into 90\% for training and 10\%  for validation. Following the standard setup, we apply random rotations for data augmentation and quantize 3D coordinates to 7-bit precision per axis.

For evaluation, we use the standard dense point clouds that are widely employed in the MPEG Common Test Conditions (CTC)~\cite{schwarz2018emerging} and the JPEG Pleno CTC~\cite{jpeg_ctc_pcc_n87037_2020}. The test dataset comprises 12 point clouds from three datasets: 8iVFB (Longdress, Loot, Redandblack, Soldier)~\cite{8i_voxelized_full_bodies_wg11m40059_wg1m74006_2017}, Owlii~(Basketball\_player, Dancer, Exercise, Model)~\cite{owlii_dynamic_human_mesh_m41658_2017}, and MVUB (Andrew, David, Phil, Sarah)~\cite{mvub_microsoft_voxelized_upper_bodies_m38673_m72012_2016}. These sequences cover diverse spatial scales and geometric characteristics, providing a representative benchmark for assessing RD performance and generalization.

%----------Implementation & Training Details
Our model is implemented in PyTorch with MinkowskiEngine \cite{choy20194d} for sparse 3D convolutions. 
For entropy coding, we adopt a fully factorized entropy model, using the reference implementation provided in CompressAI~\cite{begaint2020compressai} across all layers and channel groups to estimate the likelihoods of quantized latent symbols. For training, we use Adam with weight decay and gradient clipping, and train for 50 epochs with a batch size of 32. The learning rate is scheduled from $8\times10^{-4}$ to $2\times10^{-5}$ using a cosine decay. During training, we randomly sample a decode level $L$ at each iteration and optimize the RD loss conditioned on the selected level.

\section{Results}
\label{sec:Results}

\sisetup{
  mode = text,
  detect-weight = true,
  detect-family = true,
  input-signs = +-,
  table-number-alignment = center
}

\begin{table}[t]

\centering
\caption{Average BD-Rate and BD-PSNR (D1/D2) of TAFA-GSGC relative to PCGCv2 and G-PCC anchors on standard dense point cloud datasets.}
\label{tab:bd_rate_summary}
\vspace{2mm}
\newcommand{\tblfont}{\fontsize{8}{9}\selectfont}
\renewcommand{\arraystretch}{1.4}
\setlength{\tabcolsep}{2.5pt}
%\captionsetup{skip=4pt}
\setlength{\textfloatsep}{8pt}
\setlength{\intextsep}{8pt}
% -----------------------------------------------

{\resizebox{\columnwidth}{!}{
\begin{tabular}{|c|c|
  S[table-format=+2.2] S[table-format=+2.2]|
  S[table-format=+2.2] S[table-format=+2.2]|
  S[table-format=+2.2] S[table-format=+2.2]|
}
\hline
\multirow{2}{*}{Dataset} & \multirow{2}{*}{Metric} &
\multicolumn{2}{c|}{\textbf{PCGCv2}} &
\multicolumn{2}{c|}{G-PCC Octree} &
\multicolumn{2}{c|}{G-PCC Trisoup} \\
\cline{3-8}
& & {D1} & {D2} & {D1} & {D2} & {D1} & {D2} \\
\hline

\multirow{2}{*}{8iVFB}
& BD-Rate(\%)   & -5.43 & -5.33 & \multicolumn{1}{c}{--} & -76.22 & -31.77 & -30.88 \\
\cline{2-8}
& BD-PSNR(dB)   & +0.25 & +0.32 & +9.08 & +7.90  & +1.60  & +1.46  \\
\hline

\multirow{2}{*}{MVUB}
& BD-Rate(\%)   & -0.87 & -5.06 & -82.76 & -75.46 & \multicolumn{1}{c}{--} & \multicolumn{1}{c|}{--} \\
\cline{2-8}
& BD-PSNR(dB)   & +0.08 & +0.25 & +8.13  & +6.69  & +2.76  & +2.17  \\
\hline

\multirow{2}{*}{Owlii}
& BD-Rate(\%)   & -8.68 & -7.37 & \multicolumn{1}{c}{--} & \multicolumn{1}{c|}{--} & \multicolumn{1}{c}{--} & -30.86 \\
\cline{2-8}
& BD-PSNR(dB)   & +0.33 & +0.40 & +10.08 & +8.68  & +2.83  & +2.31  \\
\hline

\multirow{2}{*}{\textbf{Average}}
& \textbf{BD-Rate(\%)}   & \bfseries -4.99 & \bfseries -5.92 & \bfseries -82.76 & \bfseries -75.84 & \bfseries -31.77 & \bfseries -30.87 \\
\cline{2-8}
& \textbf{BD-PSNR(dB)}   & \bfseries +0.22 & \bfseries +0.32 & \bfseries +9.10  & \bfseries +7.76  & \bfseries +2.40  & \bfseries +1.98  \\
\hline
\end{tabular}}
}
\end{table}

\sisetup{
  mode = text,
  detect-weight = true,
  detect-family = true,
  input-signs = +-,
  table-number-alignment = center
}

\begin{table*}[t]
\centering 
\caption{Progressive decoding behavior of the 9-level configuration on \textit{longdress}.}
\vspace{2mm}
\label{tab:longdress_progressive_9level}

\newcommand{\tblfont}{\fontsize{8}{10}\selectfont}
\renewcommand{\arraystretch}{1.1}
\setlength{\tabcolsep}{6.0pt}                       
%\captionsetup{skip=3pt}
\setlength{\textfloatsep}{8pt}
\setlength{\intextsep}{8pt}
% -----------------------------------------------

{\tblfont
\begin{tabular}{|c|
  S[table-format=1.3]|
  S[table-format=2.3]|
  S[table-format=2.3]|
  S[table-format=+1.3]|
  S[table-format=+1.2]|
  S[table-format=+1.2]|
}
\hline
\textbf{Level} &
\textbf{Bitrate (bpp)} &
\textbf{D1 PSNR (dB)} &
\textbf{D2 PSNR (dB)} &
{$\boldsymbol{\Delta}$\textbf{Bitrate}$\uparrow$} &
{$\boldsymbol{\Delta}$\textbf{D1 PSNR}$\uparrow$} &
{$\boldsymbol{\Delta}$\textbf{D2 PSNR}$\uparrow$} \\
\hline

1 & 0.096 & 67.491 & 71.113
  & \multicolumn{1}{|c|}{--}
  & \multicolumn{1}{c|}{--}
  & \multicolumn{1}{c|}{--} \\
2 & 0.150 & 71.300 & 74.775 & +0.054 & +3.81 & +3.66 \\
3 & 0.207 & 72.328 & 75.810 & +0.057 & +1.03 & +1.04 \\
4 & 0.428 & 75.739 & 79.982 & +0.221 & +3.41 & +4.17 \\
5 & 0.509 & 76.251 & 80.596 & +0.081 & +0.51 & +0.61 \\
6 & 0.556 & 76.389 & 80.747 & +0.047 & +0.14 & +0.15 \\
7 & 0.934 & 79.197 & 83.839 & +0.378 & +2.81 & +3.09 \\
8 & 1.092 & 80.325 & 85.045 & +0.158 & +1.13 & +1.21 \\
9 & 1.233 & 80.729 & 85.456 & +0.141 & +0.40 & +0.41 \\
\hline
\end{tabular}
}
\end{table*}

\begin{figure}[t]
    \centering
    % ===== knobs you can tune =====
    %\setlength{\abovecaptionskip}{5pt}   % 图和caption之间的距离（caption在下方）
    %\setlength{\belowcaptionskip}{-15pt}  % caption和正文之间的距离（可用负值更紧凑）
    % ==============================
    \includegraphics[width=1.0\linewidth]{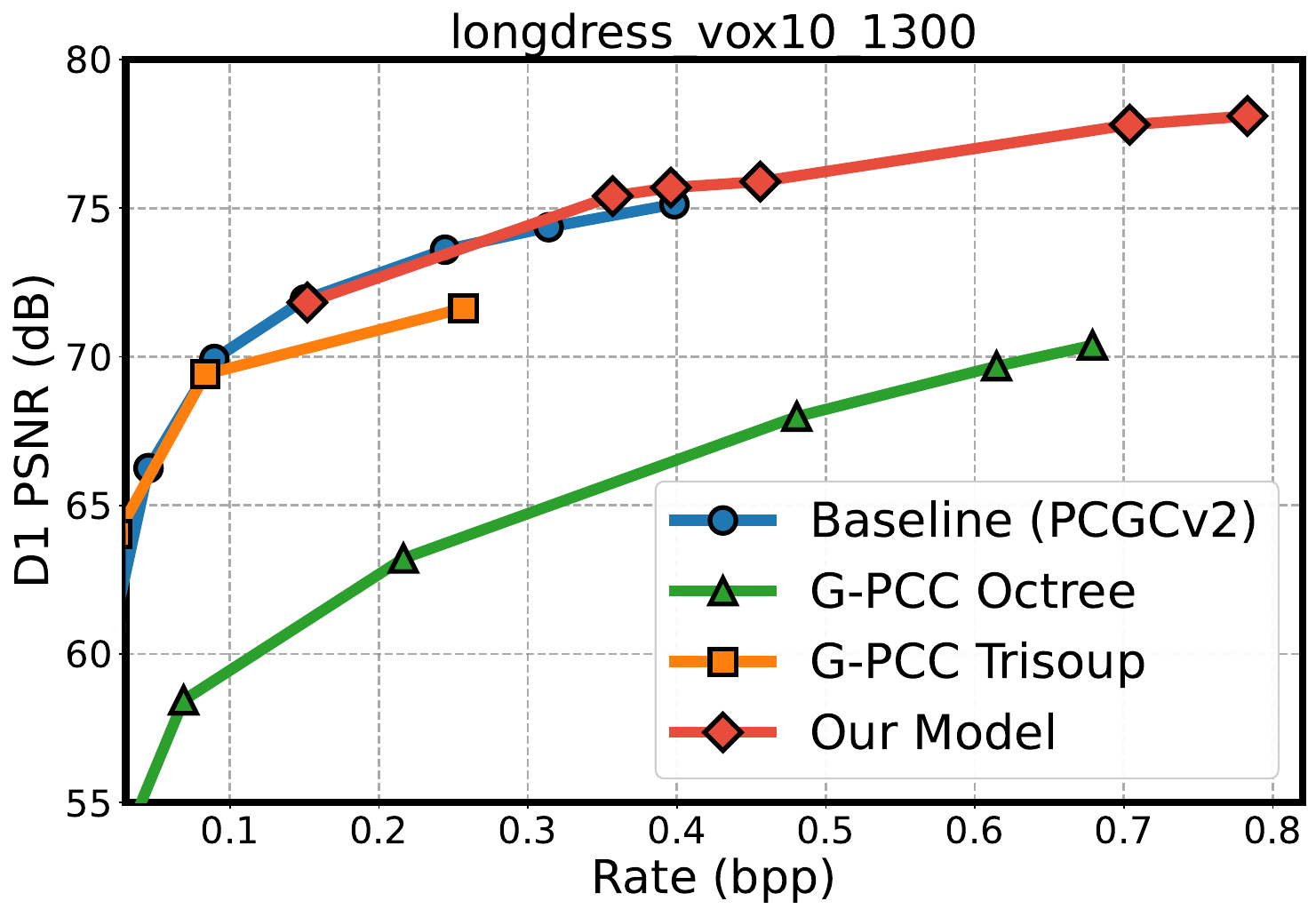} % .pdf/.png/.jpg
    \caption{RD performance on \textit{Longdress} comparing TAFA-GSGC with PCGCv2 and G-PCC anchors.}
    \label{fig:rd_curve_longdress_6level}
\end{figure}

\begin{figure*}[t]
  \centering

  % ===== spacing knobs (local to this figure) =====
              % space above the figure (tighten)
  % \captionsetup{skip=0.5pt}       % space between figure and caption
  % ===============================================

  \begin{minipage}[t]{0.4\textwidth}
    \centering
    \includegraphics[width=\linewidth]{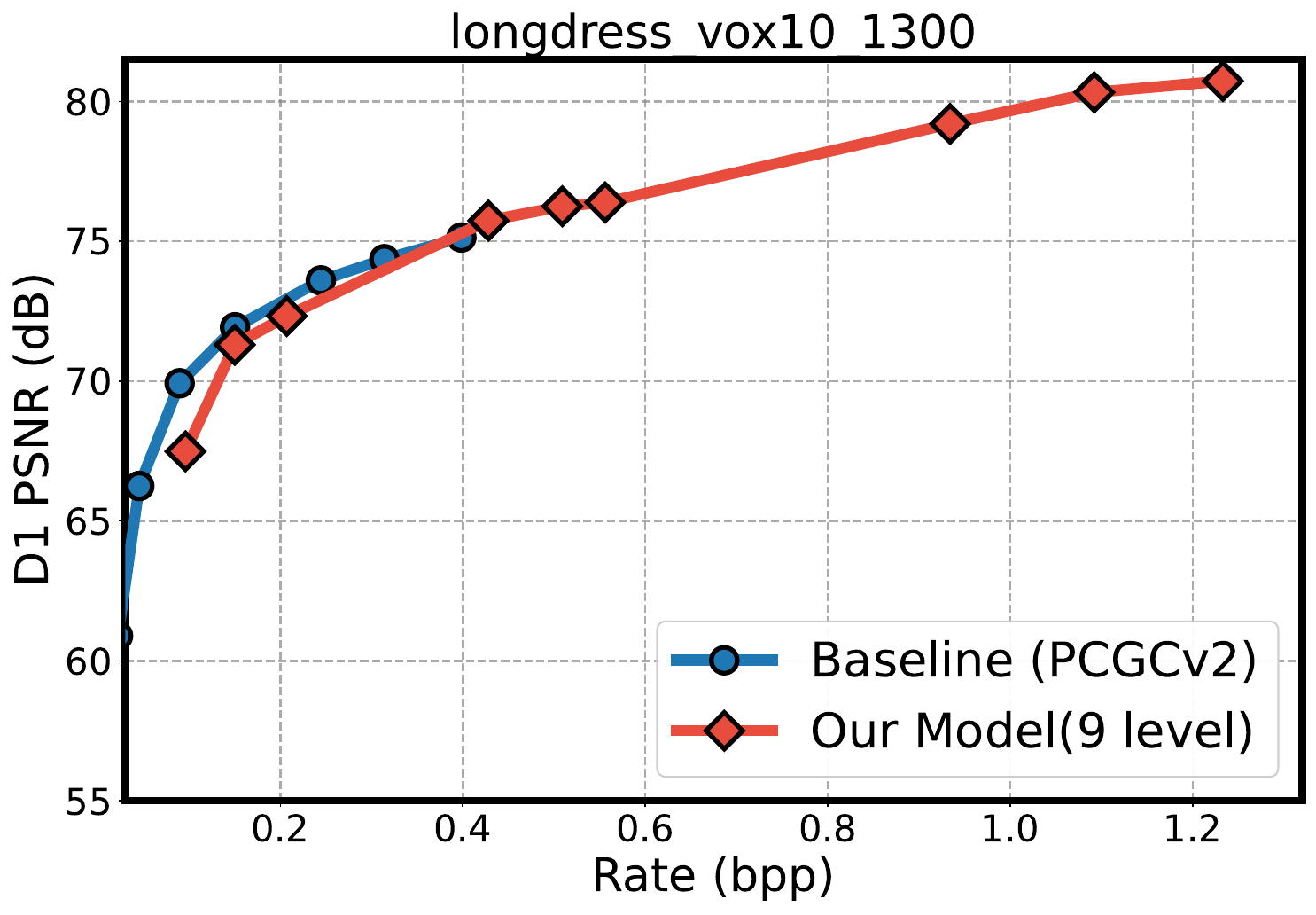}
    % \caption{D1 (p2p)}
    %\label{fig:rd_longdress_d1}
  % \end{minipage}\hfill
  \end{minipage}
  \hspace{2mm}
  \begin{minipage}[t]{0.4\textwidth}
    \centering
    \includegraphics[width=\linewidth]{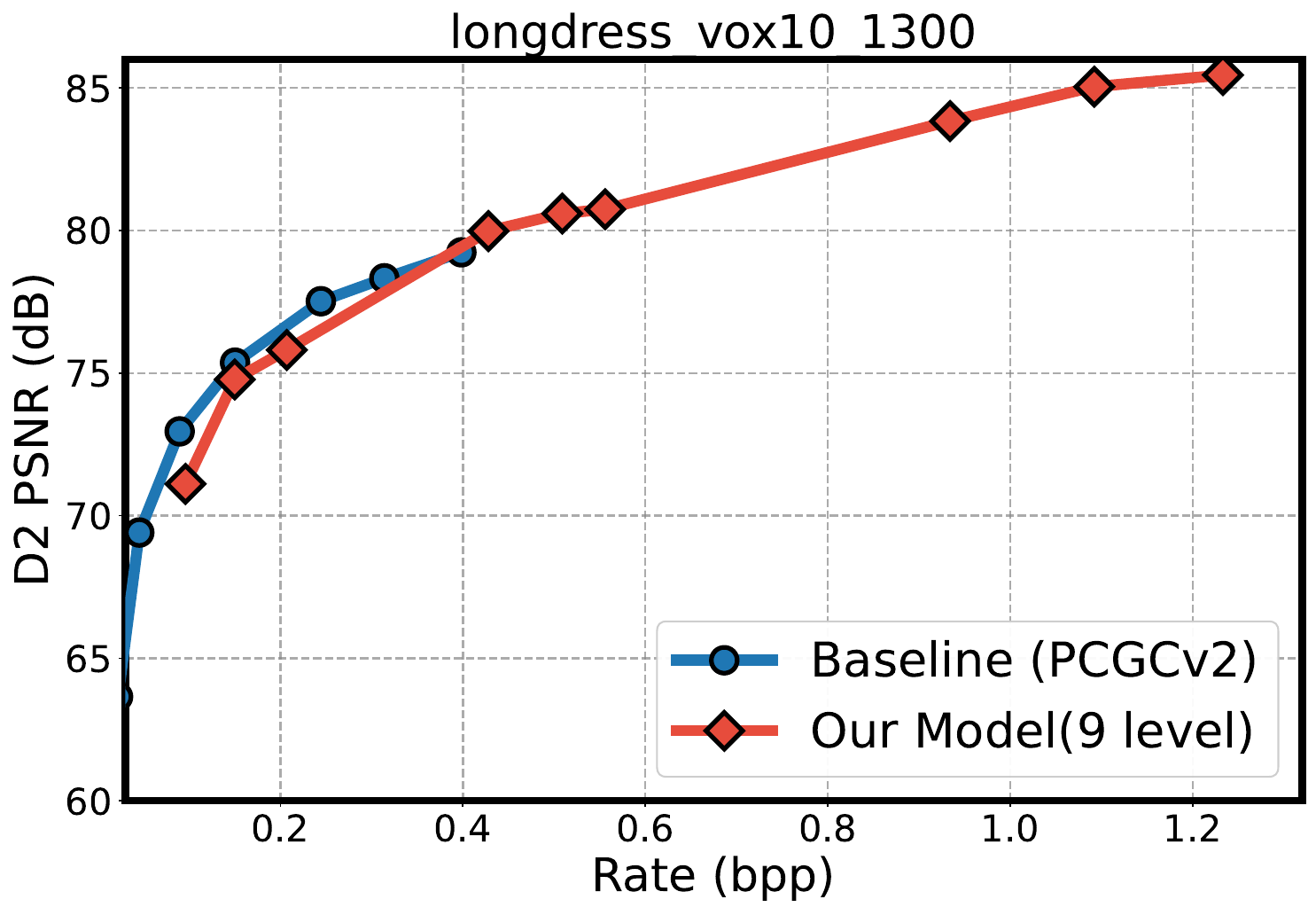}

    % \caption{D2 (p2plane)}
    %\label{fig:rd_longdress_d2}
  \end{minipage}
             % space between subfig row and main caption
  % \caption{Rate--distortion curves on \textit{Longdress} (9-level configuration).}
  \caption{RD curves on \textit{Longdress} (9-level configuration). Left: D1 (point-to-point) PSNR vs.\ bitrate. Right: D2 (point-to-plane) PSNR vs.\ bitrate.}

  \label{fig:rd_curve_longdress_9level}

\end{figure*}

We compare against the learned anchor PCGCv2 and the standardized MPEG G-PCC anchors (Octree and Trisoup) using the reference implementation TMC13-v23~\cite{mpeg_pcc_tmc13_v23} with CTC-compliant configurations. For a fair and reproducible comparison, we use the same voxelized test assets as PCGCv2.  Distortion is measured using both point-to-point distance (D1) and point-to-plane distance (D2) metrics, reported as Peak Signal-to-Noise Ratio (PSNR) in dB. The bitrate is reported in bits per input point (bpp). 

Figure~\ref{fig:rd_curve_longdress_6level} and Table~\ref{tab:bd_rate_summary} present the RD performance and BD-rate gains of our framework against PCGCv2 and G-PCC. A critical distinction of our approach is that all rate-quality points are obtained from \textit{a single scalable model} producing \textit{a single bitstream}: each quality point corresponds to a different bitstream level, without retraining or switching models. For each point, the decoder outputs a complete reconstructed point cloud at the original resolution, while the output point count is controlled by $K$, which is set according to the point count at the corresponding encoder scale. In our configuration, each layer uses an 8-channel latent representation. The channels are grouped as $(4,4)$ for the Base Layer, $(2,2,4)$ for Layer~1, and $(8)$ for Layer~2, yielding a total of 6 progressive sub-bitstreams.

As shown in Fig.~\ref{fig:rd_curve_longdress_6level}, our scalable codec matches the baseline PCGCv2 in the overlapping bitrate range, while extending the operating range to higher bitrates through progressive decoding. In the medium-to-high bitrate regime, our method consistently achieves higher reconstruction quality than PCGCv2 at comparable rates. Compared with the non-learned anchors, our method substantially outperforms both G-PCC Octree and G-PCC Trisoup over their evaluated operating points, while providing a wide bitrate coverage enabled by progressive truncation.

Table~\ref{tab:bd_rate_summary} further summarizes the average BD metrics over all sequences.
Entries marked with ``--'' indicate non-overlapping RD curves over the tested range, making BD metrics undefined. In these cases, our RD curves lie above the anchor across all evaluated bitrates, indicating clear gains.
Against G-PCC Octree, our method achieves average BD-PSNR gains of \mbox{\SI{+9.10}{\decibel}} (D1) and
\mbox{\SI{+7.76}{\decibel}} (D2), with corresponding BD-Rate savings of
\mbox{\SI{-82.76}{\percent}} and \mbox{\SI{-75.84}{\percent}}, respectively.
Compared to G-PCC Trisoup, we obtain BD-PSNR gains of \mbox{\SI{+2.40}{\decibel}} (D1) and
\mbox{\SI{+1.98}{\decibel}} (D2), corresponding to BD-Rate savings of
\mbox{\SI{-31.77}{\percent}} and \mbox{\SI{-30.87}{\percent}}.
Relative to the baseline PCGCv2, our method maintains essentially the same RD performance on average, with BD-Rate savings of \mbox{\SI{-4.99}{\percent}} (D1) and \mbox{\SI{-5.92}{\percent}} (D2), and BD-PSNR gains of \mbox{\SI{+0.22}{\decibel}} and \mbox{\SI{+0.32}{\decibel}}.

\subsection{Scalability Analysis}
\label{ssec:ScalabilityAnalysis}

In addition to the main configuration, we report a 9-level configuration on \textit{longdress} in D1 and D2 to explicitly illustrate the progressive decoding behavior of our scalable bitstream, as shown in Fig.~\ref{fig:rd_curve_longdress_9level} and Table~\ref{tab:longdress_progressive_9level}. In this setting, each layer uses an 8-channel latent representation and applies the same channel grouping $(2,2,4)$, resulting in $9$ levels. Table~\ref{tab:longdress_progressive_9level} lists the resulting bitrate-quality points and the corresponding per-level bitrate and D1/D2 PSNR changes. As more sub-bitstreams are decoded in order, the reconstruction quality improves monotonically, demonstrating fine-grained rate adaptation with predictable quality gains.

\section{Conclusion}
\label{sec:print}

We propose TAFA-GSGC, a scalable learned point cloud geometry codec built on the PCGCv2 backbone. A single trained model enables multi-level decoding by progressively truncating a single bitstream. Experiments on standard dense point cloud datasets show that TAFA-GSGC substantially outperforms G-PCC anchors and achieves comparable or slightly better RD performance than PCGCv2 under a fair comparison. To our knowledge, TAFA-GSGC is the first learned geometry codec to support up to \emph{9} decodable levels within a single model while maintaining strong reconstruction quality. The number of decodable levels is determined by the channel-group configuration, allowing straightforward extension to more levels without changing the overall architecture, and our modular scalability components can be integrated into other learned geometry backbones. Future work will incorporate our design into stronger codecs and extend it to joint geometry--attribute coding.% and temporal scalability.

\vfill\pagebreak

\bibliographystyle{IEEEbib}
\bibliography{strings,refs}

\end{document}

%% file: overview.tex
\begin{tikzpicture}[
    node distance=1.5cm and 1.5cm,
    % --- GENERAL STYLING ---
    font=\sffamily\large,
    thick,
    % --- COMPONENT STYLES ---
    base_block/.style={
        rectangle, 
        draw=black!80, 
        minimum width=1.5cm, 
        minimum height=0.8cm, 
        align=center,
        rounded corners=3pt
    },
    encoder_block/.style={
        base_block,
        fill=blue!10,
        draw=blue!60!black,
        line width=1pt
    },
    decoder_block/.style={
        base_block,
        fill=orange!10,
        draw=orange!60!black,
        line width=1pt
    },
    prior_block/.style={ 
        base_block,
        minimum width=1.5cm, 
        fill=teal!10,
        draw=teal!60!black,
        line width=1pt
    },
    novelty_layer/.style={
        rectangle,
        draw=purple!70!black,
        fill=purple!15,
        line width=1.5pt,
        minimum width=1.5cm,
        minimum height=0.7cm,
        align=center,
        rounded corners=0pt
    },
    entropy_coder/.style={
        rectangle,
        rounded corners=6pt,
        draw=lime!60!black,
        fill=lime!20,
        line width=1pt,
        minimum width=1.2cm,
        minimum height=0.7cm,
        align=center
    },
    conv_layer/.style={
        rectangle,
        draw=gray!60!black,
        fill=gray!10,
        minimum width=1.5cm,
        minimum height=0.7cm,
        align=center,
        rounded corners=0pt,
        line width=1pt
    },
    small_sq/.style={
        draw,
        fill=white,
        align=center,
        font=\sffamily\large,
        minimum size=0.8cm 
    },
    bit_sq/.style={draw, minimum size=1.2em}, 
    container_box/.style={
        draw=gray!70, fill=containergray!40, very thick, 
        rounded corners=20pt, inner sep=15pt
    },
    connector/.style={
        ->,
        >={Latex[length=2.5mm, width=1.5mm]},
        draw=black!70,
        line width=1pt
    },
    % --- NEW: Single Node Split Stream ---
    split_stream/.style={
        rectangle,
        draw=black!70,
        minimum width=2.4em, 
        minimum height=1.2em,
        inner sep=0pt,
        fill=white, 
        path picture={
            \fill[bitgray] (path picture bounding box.south west) rectangle ($(path picture bounding box.north west)!0.5!(path picture bounding box.north east)$);
            \draw[black!70] ($(path picture bounding box.south west)!0.5!(path picture bounding box.south east)$) -- ($(path picture bounding box.north west)!0.5!(path picture bounding box.north east)$);
        }
    },
    % --- LEGEND STYLES ---
    legend_icon/.style={
        minimum width=1.5cm,
        minimum height=1cm, 
        text width=1cm, % <--- KEY FIX: Forces text box to fill the node width
        align=center,     % <--- KEY FIX: Centers text within that box
        font=\sffamily\large\bfseries, 
        anchor=center,
        inner sep=0pt     % Removes internal padding to ensure exact centering
    },
    legend_text/.style={anchor=west, inner sep=5pt}
]

% --- START MAIN DIAGRAM SCOPE ---
\begin{scope}[local bounding box=mainDiagram]
    \node (input) {$\bm{X}_\text{in}$};

    \node[encoder_block, right=2.3cm of input] (ds0) {DS0};
    \node[encoder_block, right=5cm of ds0] (ds1) {DS1};
    \node[encoder_block, right=5cm of ds1] (ds2) {DS2};

    \node[conv_layer, below right = -0cm and 0.5cm of ds2] (lp) {SConv};

    \node[small_sq, below=0.35cm of lp] (base_Q) {Q};

    \node[entropy_coder, below=0.35cm of base_Q] (base_AE) {AE};
    \node[entropy_coder, right=1.5cm of base_AE] (OE) {OE};

    \node[split_stream, below=0.35cm of base_AE.south] (bae_mid) {};
    \node[split_stream, below left=0.35cm and 0.7cm of base_AE.south] (bae_left) {};
    \node[split_stream, below right=0.35cm and 0.7cm of base_AE.south] (bae_right) {};

    \node[split_stream, below=0.35cm of OE.south] (oe_mid) {};

    \node[entropy_coder, below=0.35cm of oe_mid] (OD) {OD};
    \node[entropy_coder, below=0.35cm of bae_mid] (base_AD) {AD};

    \coordinate[xshift=1.75cm] (enhancement1) at (ds1); 

    \node[novelty_layer] (tafa1) at (enhancement1 |- lp) {TAFA};
    \node[small_sq, below=0.35cm of tafa1] (enhancement1_Q) {Q};
    \node[entropy_coder, below=0.35cm of enhancement1_Q] (enhancement1_AE) {AE};

    \node[split_stream, below=0.35cm of enhancement1_AE.south] (e1ae_mid) {};
    \node[split_stream, below left=0.35cm and 0.7cm of enhancement1_AE.south] (e1ae_left) {};
    \node[split_stream, below right=0.35cm and 0.7cm of enhancement1_AE.south] (e1ae_right) {};

    \node[entropy_coder, below=0.35cm of e1ae_mid] (enhancement1_AD) {AD};

    \node[small_sq, circle, outer sep=0pt, inner sep=2pt, below=0.35cm of enhancement1_AD, minimum width=0cm, minimum height=0cm] (cat1) {C};
    \node[conv_layer, below=0.35cm of cat1] (fusion1) {SConv};

    \coordinate[xshift=1.75cm] (enhancement2) at (ds0);

    \node[novelty_layer] (tafa2) at (enhancement2 |- lp) {TAFA};
    \node[small_sq, below=0.35cm of tafa2] (enhancement2_Q) {Q};
    \node[entropy_coder, below=0.35cm of enhancement2_Q] (enhancement2_AE) {AE};

    \node[split_stream, below=0.35cm of enhancement2_AE.south] (e2ae_mid) {};
    \node[split_stream, below left=0.35cm and 0.7cm of enhancement2_AE.south] (e2ae_left) {};
    \node[split_stream, below right=0.35cm and 0.7cm of enhancement2_AE.south] (e2ae_right) {};

    \node[entropy_coder, below=0.35cm of e2ae_mid] (enhancement2_AD) {AD};

    \node[small_sq, circle, outer sep=0pt, inner sep=2pt, below=0.35cm of enhancement2_AD, minimum width=0cm, minimum height=0cm] (cat2) {C};
    \node[conv_layer, below=0.35cm of cat2] (fusion2) {SConv};

    \node[decoder_block] (us2) at ([yshift=-1cm]fusion1.south -| ds2) {US2};
    \node[decoder_block] (us1) at (ds1 |- us2) {US1};
    \node[decoder_block] (us0) at (ds0 |- us1) {US0};

    \node[conv_layer, left=0.3cm of us2, minimum width=1.0cm] (cls_2) {Prune};
    \node[conv_layer, left=0.3cm of us1, minimum width=1.0cm] (cls_1) {Prune};
    \node[conv_layer, left=0.3cm of us0, minimum width=1.0cm] (cls_0) {Prune};

    \node (output) at (input |- us0) {$\bm{X}_\text{out}$};

    \draw[connector] (input) -- (ds0);
    % \draw[connector] (ds0) -- (ds1);
    \draw[connector] (ds0) -- node[pos=0.3, above] {$s{=}0$} (ds1);
    \draw[connector] (ds2) -| node[pos=0.78, right, yshift=-0pt, font=\small] {$\bm{F}_{2}$} (lp);
    \draw[connector] (ds2) -| node[pos=0.78, right, yshift=35pt, font=\small] {$\bm{C}_{2}$} (OE);

    % \draw[connector] (ds1) -- (ds2);
    % \draw[connector] (ds1) -- node[midway, above] {$s{=}1$} (ds2);
    \draw[connector] (ds1) -- 
  node[midway, above] {$s{=}1$}
  node[pos=0.25, above, font=\small] {$\bm{E}_1$}
  (ds2);

    % \draw[connector] (ds2) -| (lp);
    \draw[connector] (ds2) -| node[midway, above] {$s{=}2$} (lp);
    \draw[connector] (lp) -- (base_Q);
    \draw[connector] (base_Q) -- (base_AE);

    \draw[connector] (ds2) -| (OE);
    \draw[connector] (OE) -- (oe_mid);
    \draw[connector] (oe_mid) -- (OD);
     
    \draw[connector] (base_AE) -- (bae_left);
    \draw[connector] (base_AE) -- (bae_mid);
    \draw[connector] (base_AE) -- (bae_right);

    \draw[connector] (bae_left) -- (base_AD);
    \draw[connector] (bae_mid) -- (base_AD);
    \draw[connector] (bae_right) -- (base_AD);

    \draw[connector] (base_AD) -- (us2 -| base_AD);
    \draw[connector] (OD) |- (us2);

    \draw[connector] (enhancement1) -- (tafa1);
    \draw[connector] (tafa1) -- (enhancement1_Q);
    % \draw[connector] (tafa1) -- node[midway, right, font=\small] {$\bm{E}'_{1}$} (enhancement1_Q);

    \draw[connector] (enhancement1_Q) -- (enhancement1_AE);

    \draw[connector] (enhancement1_AE) -- (e1ae_left);
    \draw[connector] (enhancement1_AE) -- (e1ae_mid);
    \draw[connector] (enhancement1_AE) -- (e1ae_right);

    \draw[connector] (e1ae_left) -- (enhancement1_AD);
    \draw[connector] (e1ae_mid) -- (enhancement1_AD);
    \draw[connector] (e1ae_right) -- (enhancement1_AD);

    \draw[connector] (enhancement1_AD) -- (cat1);
    \draw[connector] (cat1) -- (fusion1);

    \draw[connector] (cls_2.west) -- ++(-0.25, 0) |- (cat1);
    \draw[connector] (cls_2.west) -- ++(-0.25, 0) |- (tafa1);

    \coordinate (pivot1) at (fusion1 |- us2);

    \node[circle, fill=black, inner sep=1.5pt] (contact_fusion1) at ([yshift=0.75cm] pivot1) {};
    \node[circle, fill=black, inner sep=1.5pt] (contact_us2) at ([xshift=0.433cm] pivot1) {};
    \node[circle, fill=black, inner sep=1.5pt] (contact_us1_east) at ([xshift=-0.433cm] pivot1) {};

    \draw[connector, -] (fusion1) -- (contact_fusion1);
    \draw[connector] (contact_us1_east) -- (us1);
    \draw[connector, -] (cls_2) -- (contact_us2);

    \draw[connector] (us2) -- (cls_2);
    % \node[font=\sffamily\normalsize, anchor=north] at ([xshift=60pt,yshift=8pt]us2.south) {$s{=}2$};
    \node[anchor=north] at ([xshift=40pt,yshift=10pt]us2.south) {$s{=}2$};

  \node[font=\sffamily\scriptsize, anchor=south] 
  at ([xshift=7pt,yshift=50pt]cls_2.north) {$\hat{\bm{X}}_{1}=\{\hat{\bm{C}}_{1},\hat{\bm{F}}_{1}\}$};

\node[font=\sffamily\scriptsize, anchor=north]
  at ([xshift=30pt, yshift=-7pt]fusion1.south)
  {$\tilde{\bm{X}}_{1}=\{\hat{\bm{C}}_{1},\tilde{\bm{F}}_{1}\}$};

  % --- layer labels under DS blocks ---
\node[font=\sffamily\small\bfseries, anchor=north] 
  at ([yshift=-10pt]ds0.south) {Layer 2};

\node[font=\sffamily\small\bfseries, anchor=north] 
  at ([yshift=-10pt]ds1.south) {Layer 1};

  \node[font=\sffamily\small\bfseries, anchor=north]
  at ([yshift=-10pt]ds2.south) {Base Layer};

    \draw[connector] (contact_fusion1) -- coordinate(arc1) (contact_us1_east);
    \draw[dashed, <->, line width=1pt] ($(arc1) + (-0.2, 0.3)$) arc (60:25:1.6);

    \draw[connector] (enhancement2) -- (tafa2);
    \draw[connector] (tafa2) -- (enhancement2_Q);
    \draw[connector] (enhancement2_Q) -- (enhancement2_AE);

    \draw[connector] (enhancement2_AE) -- (e2ae_left);
    \draw[connector] (enhancement2_AE) -- (e2ae_mid);
    \draw[connector] (enhancement2_AE) -- (e2ae_right);

    \draw[connector] (e2ae_left) -- (enhancement2_AD);
    \draw[connector] (e2ae_mid) -- (enhancement2_AD);
    \draw[connector] (e2ae_right) -- (enhancement2_AD);

    \draw[connector] (enhancement2_AD) -- (cat2);
    \draw[connector] (cat2) -- (fusion2);

    \draw[connector] (cls_1.west) -- ++(-0.25, 0) |- (cat2);
    \draw[connector] (cls_1.west) -- ++(-0.25, 0) |- (tafa2);

    \coordinate (pivot2) at (fusion2 |- us1);

    \node[circle, fill=black, inner sep=1.5pt] (contact_fusion2) at ([yshift=0.75cm] pivot2) {};
    \node[circle, fill=black, inner sep=1.5pt] (contact_us1_west) at ([xshift=0.433cm] pivot2) {};
    \node[circle, fill=black, inner sep=1.5pt] (contact_us0) at ([xshift=-0.433cm] pivot2) {};

    \draw[connector, -] (fusion2) -- (contact_fusion2);
    \draw[connector] (contact_us0) -- (us0);
    \draw[connector, -] (cls_1) -- (contact_us1_west);

    \draw[connector] (us1) -- (cls_1);
    % \node[font=\sffamily\normalsize, anchor=north] at ([xshift=90pt,yshift=8pt]us1.south) {$s{=}1$};
    \node[anchor=north] at ([xshift=90pt,yshift=10pt]us1.south) {$s{=}1$};

    \draw[connector] (contact_fusion2) -- coordinate(arc2) (contact_us0);
    \draw[dashed, <->, line width=1pt] ($(arc2) + (-0.2, 0.3)$) arc (60:25:1.6);

    \draw[connector] (us0) -- (cls_0);
    % \node[font=\sffamily\normalsize, anchor=north] at ([xshift=90pt, yshift=8pt]us0.south) {$s{=}0$};
    \node[anchor=north] at ([xshift=80pt, yshift=10pt]us0.south) {$s{=}0$};

    \draw[connector] (cls_0) -- (output);
\end{scope}
% --- END MAIN DIAGRAM SCOPE ---

% --- 2-COLUMN LEGEND MATRIX ---
%\matrix [
%    matrix of nodes,
%    nodes in empty cells,
%    draw=black!30,      
%    fill=white,         
%    rounded corners=5pt,
%    inner sep=10pt,
%    row sep=8pt,
%    column sep=5pt, 
%    anchor=north west,  
%    right=0.5cm of mainDiagram,
%    ampersand replacement=\&, 
    % We define a spacer column (Column 3) to separate the two logical groups
%    column 3/.style={minimum width=1cm} 
%] (legend) {
%    |[encoder_block, legend_icon]| {DS} \& |[legend_text]| {\ml{Downsampling\\Block}} \&
%    |[decoder_block, legend_icon]| {US} \& |[legend_text]| {\ml{Upsampling\\Block}} \\
%    |[entropy_coder, legend_icon]| {AE} \& |[legend_text]| {\ml{Arithmetic\\Encoder}} \&
%    |[entropy_coder, legend_icon]| {AD} \& |[legend_text]| {\ml{Arithmetic\\Decoder}} \\ 
%    |[entropy_coder, legend_icon]| {OE} \& |[legend_text]| {\ml{Octree\\Encoder}} \&
%    |[entropy_coder, legend_icon]| {OD} \& |[legend_text]| {\ml{Octree\\Decoder}} \\
%    |[conv_layer, legend_icon]| {CLS}   \& |[legend_text]| {Classifier} \& 
%    |[conv_layer, legend_icon]| {Conv}  \& |[legend_text]| {\ml{Sparse\\Convolution}} \\
%    |[small_sq, legend_icon]| {Q}       \& |[legend_text]| {Quantization} \& 
%    |[small_sq, circle, inner sep=1pt, font=\sffamily\small\bfseries, legend_icon, minimum width=0.5cm, minimum height=0.5cm, text width=0.5cm]| {C} \& |[legend_text]| {Concatenation} \\
%};

% Add a title to the legend
%\node[above=5pt, font=\sffamily\large\bfseries] at (legend.north) {Legend};

\end{tikzpicture}

%% file: tafa.tex
\begin{tikzpicture}[
    node distance=1.5cm and 1.5cm,
    % --- GENERAL STYLING ---
    font=\sffamily\large,
    thick,
    % --- COMPONENT STYLES ---
    base_block/.style={
        rectangle, 
        draw=black!80, 
        minimum width=2.5cm, 
        minimum height=1.2cm, 
        align=center,
        rounded corners=3pt
    },
    novelty_layer/.style={
        rectangle,
        % draw=purple!70!black,
        % fill=purple!15,
        draw=orange!80!black,
        fill=orange!15,
        line width=1.5pt,
        minimum width=2.3cm,
        minimum height=0.7cm,
        align=center,
        rounded corners=0pt
    },
    activation/.style={
        rectangle,
        draw=cyan!70!black,
        fill=cyan!10,
        minimum width=0.75cm,
        minimum height=0.6cm,
        align=center,
        rounded corners=2pt,
        font=\sffamily\footnotesize % Optional: slightly smaller text
    },
    conv_layer/.style={
        rectangle,
        draw=gray!60!black,
        fill=gray!10,
        minimum width=2.3cm, 
        minimum height=0.7cm,
        align=center,
        rounded corners=0pt,
        line width=1pt
    },
    connector/.style={
        ->,
        >={Latex[length=2.5mm, width=1.5mm]},
        draw=black!70,
        line width=1pt
    },
    small_sq/.style={
        draw,
        fill=white,
        minimum size=2em,
        align=center,
        font=\sffamily\large
    },
]

% $\hat{\bm{X}}_{s}=\{\hat{\bm{C}}_{s},\,\hat{\bm{F}}_{s}\}$

    \node (input) {$\bm{E}_{s} = \{{\bm{C}}_{s},\,{\bm{F}}_{s}\} $};

    \coordinate[below=0.85cm of input] (upper);
    \node[novelty_layer, below=1.25cm of input] (acc) {Target-Aligned\\SConv};

    \node[activation, below=0.35cm of acc] (relu) {RELU};

    \node[small_sq, circle, inner sep=2pt, below=0.35cm of relu] (cat) {C};

    \node[conv_layer, below=0.35cm of cat] (final) {SConv};

    % \node[below right=1cm and 0.5cm of final, anchor=center] (condition) {$\bm{\hat{y}}$};
%    \node[below right=1.5cm and 0.5cm of final, anchor=west, font=\sffamily\scriptsize] (condition) {$\hat{\bm{X}}_{s}=\{\hat{\bm{C}}_{s},\,\hat{\bm{F}}_{s}\}$};

    \node[below=1.5cm of final, anchor=center] (output) {$\bm{y}_{\mathrm{enh}}^{(s)}$};
    \coordinate[above right=1cm and 1.2cm of final] (lower);

    \draw[connector] (input) -- (acc);
    \draw[connector] (acc) -- (relu);

    % \draw[connector] (relu) -- (cat);
    \draw[connector] (relu) -- node[midway, right, yshift=-1pt,  font=\sffamily\normalsize] {$\bm{E}'_{s}$} (cat);

    \draw[connector] (cat) -- (final);
    \draw[connector] (final) -- (output);

    % \draw[connector] (condition) |- (cat);
   % \draw[connector] (condition) |- node[pos=0.75, above=-1pt, font=\sffamily\scriptsize] {$\hat{\bm{F}}_{s}$} (cat);

    %\draw[connector] (condition) |- (acc);

%\node[font=\sffamily\scriptsize, anchor=south] 
%  at ([xshift=20pt, yshift=132pt] acc.east |- condition) {$\hat{\bm{C}}_{s}$};

% % ---------- Outer box for the whole TAFA module ----------
\begin{scope}[on background layer]
    \node[
      draw=purple!70!black,
      fill=purple!15,
      rounded corners=3pt,
      line width=2pt,
      inner sep=10pt,
      fit=(upper)(acc)(relu)(cat)(final)(lower),
      label={[font=\sffamily\bfseries\large, anchor=south east]north east:TAFA}
    ] (tafa_box) {};
\end{scope}
% % ---------------------------------------------------------

    \node[below right=1.5cm and 0.5cm of final, anchor=west] (condition) {$\hat{\bm{X}}_{s}=\{\hat{\bm{C}}_{s},\,\hat{\bm{F}}_{s}\}$};
    
    \draw[connector] (condition.north) |- (tafa_box.east) --  ++(-0.5, 0) |- node[pos=0.62, above=-1pt, font=\sffamily\normalsize]{$\hat{\bm{C}}_{s}$} (acc);
    \draw[connector, font=\sffamily\normalsize] (condition.north) |- (tafa_box.east) --  ++(-0.5, 0) |- node[pos=0.56, below=-1pt]{$\hat{\bm{F}}_{s}$}(cat);

\end{tikzpicture}